%% file: arxiv.tex
\documentclass{article} % For LaTeX2e
\usepackage{iclr2027_conference,times}

\input{math_commands.tex}

\usepackage{hyperref}
\usepackage{url}
\usepackage{graphicx}
\usepackage{booktabs}
\usepackage{multirow}
\usepackage{enumitem}

\title{Teach Me After You Learn It: Lookahead Self-Distillation for RLVR}

\author{
Zepeng Zhai
\quad
Meilin Chen\thanks{Corresponding author.} \\
Xiaohongshu Inc. \\
\texttt{\{zhaizepeng,chenmeilin2\}@xiaohongshu.com}
}

\iclrfinalcopy % Keep author names visible in the arXiv preprint.
\begin{document}

\maketitle
\lhead{} % Remove the template's premature "Published at ICLR 2027" claim.

\noindent\textbf{TL;DR:} The self-distillation teacher first learns from \emph{other} on-policy rollouts: a virtual RL update on one half of each rollout group scores the other half, and the agreement between that signal and the GRPO advantage modulates the update.
\medskip

\begin{abstract}
On-policy self-distillation (OPSD) has emerged as a popular training paradigm for LLMs, providing dense, token-level supervision by aligning a model’s own distribution with the distribution induced by privileged information. However, existing OPSD methods derive supervision solely from privileged information of the same instance, leaving inter-instance supervision largely unexplored. 
We propose \textbf{L}ookah\textbf{EA}d self-\textbf{D}istillation (\textbf{LEAD}), which introduces \emph{inter-instance} distillation by allowing the teacher to first learn from other on-policy rollouts before teaching the student. LEAD constructs lookahead teachers through virtual RL updates on disjoint subsets of on-policy rollouts, and uses each teacher to provide cross-group token-level supervision for the other subset. To reconcile the complementary yet potentially conflicting signals from OPSD and RLVR, we introduce \emph{harmonious modulation}, which adaptively modulates the GRPO advantage based on the agreement between the two signals. Extensive experiments on scientific reasoning and interactive-agent benchmarks show that LEAD achieves the best in-domain average performance across model scales while maintaining out-of-domain generalization comparable to GRPO.
\end{abstract}

\section{Introduction}
Reinforcement Learning with Verifiable Rewards (RLVR) has emerged as an effective paradigm for improving the reasoning capabilities of large language models (LLMs), where each trajectory receives only a single scalar signal determined by the outcome from the environment. 
On-policy distillation (OPD)~\citep{agarwal2024policy,gu2024minillm} complements this by leveraging a stronger teacher model to provide dense, token-level logits as learning signals along the student’s trajectories, enriching the trajectory-level supervision to the token level and thereby achieving faster convergence. Recent work has shown that OPD from advanced teachers can match or even outperform RLVR~\citep{lu2025onpolicy,li2026rethinking}, establishing it as an equally compelling paradigm.

Despite its effectiveness, OPD introduces considerable computational overhead due to relying on a separate teacher model, which is typically substantially larger than the student. In addition, OPD performs token-level distribution matching over a shared vocabulary, requiring the teacher and student to adopt the same vocabulary spaces. This architectural constraint further limits the scalability and flexibility of OPD in practice. On-policy self-distillation (OPSD)~\citep{zhao2026self,hubotter2026reinforcement} provides a more lightweight alternative by dispensing with an external teacher altogether. Instead, a single model plays both roles: it acts as the teacher when conditioned on privileged information, such as a verified reasoning trace or environmental feedback, and as the student when given only the query.

Despite these advantages, recent studies have revealed an inherent limitation of naive OPSD: the privileged teacher can induce undesirable shifts in the student's output style and cause leakage of privileged information, ultimately leading to degraded out-of-domain generalization~\citep{yang2026self, pan2026rlcsd}. Motivated by this, recent approaches seek to combine the benefits of OPSD with the robust optimization signal provided by RLVR. RLSD~\citep{yang2026self} uses the RLVR objective to determine the optimization direction, while treating the self-distillation signal as a modulation of the update magnitude. RLCSD~\citep{pan2026rlcsd} further improves the quality of privileged self-distillation signals by contrasting correct and incorrect privileged contexts, thereby suppressing task-irrelevant components such as privilege-induced style shifts. Refer to Appendix~\ref{app:related} for detailed analysis of related works.

Nevertheless, these approaches still follow the prior OPSD paradigm, where the supervision signal for the \(i\)-th query \(x_i\) in dataset \(D\) is derived from
$
e_{i,t}
=
\log \pi_{\theta}^{T}\left(y_{i,t}\mid x_i,y_{i,<t},r_i\right)
-
\log \pi_{\theta}^{S}\left(y_{i,t}\mid x_i,y_{i,<t}\right),
$
where \(y_i\) is a student-sampled rollout, \(T\) and \(S\) denote the same model under different conditioning, and \(r_i\) represents privileged information associated with query \(x_i\). In this paradigm, the supervision signal for \(x_i\) is solely derived from privileged information of the current instance \(x_i\), without explicitly considering others in dataset \(D\). We refer to this paradigm as \emph{intra-instance} distillation.
Therefore, an important question arises: 

\textbf{\emph{Can the supervision signal for an instance come not only from privileged information about the instance itself, but also from others?}}

Building on this idea, we propose \textbf{L}ookah\textbf{EA}d self-\textbf{D}istillation (\textbf{LEAD}), which introduces \emph{inter-instance} distillation by allowing the teacher to first learn from other on-policy rollouts before teaching the student. Specifically, to avoid the \emph{supervision entanglement} during inter-instance distillation, we split the on-policy rollouts into two groups and perform a \emph{Lookahead} update on each group independently, yielding two lookahead teachers. Each lookahead teacher then evaluates the rollouts from the other group, forming a \emph{Cross Scoring} mechanism that transfers knowledge acquired from one group to another. Moreover, existing approaches typically treat the OPSD and RLVR supervision signals independently, overlooking that they may provide complementary yet potentially conflicting guidance. To reconcile them, we introduce \emph{Harmonious Modulation}, which adaptively modulates the GRPO advantage according to the agreement between the two signals. 

In contrast to prior OPSD, where the supervision signal for \(x_i\) is solely obtained by conditioning the model on privileged information specific to \(x_i\), LEAD derives supervision from a teacher that has first learned from other instances. Taking advantage of this, the resulting signal provides a more effective and stable optimization direction.
Overall, our contributions are threefold:
\begin{itemize}[leftmargin=*, itemsep=2pt, topsep=2pt, parsep=0pt]
\item We propose \textbf{L}ookah\textbf{EA}d self-\textbf{D}istillation (\textbf{LEAD}), which extends self-distillation from intra-instance to inter-instance distillation, enabling the teacher to acquire information from other on-policy rollouts before providing supervision to the student.
\item We introduce \emph{Cross Scoring} to prevent supervision entanglement by explicitly transferring information across disjoint groups of on-policy rollouts, and \emph{Harmonious Modulation} to reconcile the complementary yet potentially conflicting OPSD and RLVR signals.
\item Extensive experiments across the scientific-reasoning and interactive-agent suites demonstrate LEAD achieves the best in-domain average performance across model scales while maintaining out-of-domain generalization comparable to GRPO.
\end{itemize}

\section{Preliminaries}
\label{sec:prelim}
 
\noindent \textbf{GRPO.}
Consider a language model $\pi_\theta$ trained to solve reasoning tasks.
Given a question $x$, the model generates a response $y = (y_1, \dots, y_T)$ autoregressively.
In the RLVR setting, a verifier provides a binary reward $R(x, y) \in \{0, 1\}$ indicating whether the response is correct.
Group Relative Policy Optimization (GRPO)~\citep{shao2024deepseekmath} samples a group of $G$ responses $\{y_1, \dots, y_G\}$ from the current policy for each question $x$, and computes a sequence-level advantage for each response relative to the group:
\begin{equation}
    A_i = \frac{R(x, y_i) - \mu_G}{\sigma_G},
\end{equation}
where $\mu_G$ and $\sigma_G$ are the mean and standard deviation of the rewards within the group.
The policy is then updated via a clipped surrogate objective:
\begin{equation}
    \mathcal{L}_{\text{GRPO}}(\theta) = \mathbb{E}\left[\frac{1}{G}\sum_{i=1}^{G}\frac{1}{|y_i|}\sum_{t=1}^{|y_i|} \min\!\left(\rho_i A_i,\; \text{clip}\!\left(\rho_i, 1-\epsilon, 1+\epsilon\right) A_i\right)\right],
    \label{eq:grpo}
\end{equation}
where $\rho_i = \pi_\theta(y_{i,t} \mid x, y_{i,<t}) / \pi_{\theta_{\text{old}}}(y_{i,t} \mid x, y_{i,<t})$ is the importance sampling ratio between the current and old policies.
A key limitation of GRPO is that all tokens within a response share the same advantage $A_i$, as the reward signal is provided only at the sequence level.

\noindent \textbf{OPSD.}
On-Policy Distillation (OPD)~\citep{agarwal2024policy} addresses the sparse reward problem by having the student $\pi_S$ sample its own trajectories while a separate, typically larger, teacher model $\pi_{T}$ provides dense token-level supervision.
While effective, maintaining a distinct teacher model throughout training introduces significant computational overhead.
On-Policy Self-Distillation (OPSD)~\citep{zhao2026self,hubotter2026reinforcement} eliminates this requirement by instantiating both roles from a single model $\pi_\theta$: the teacher gains its informational advantage not from greater capacity, but from conditioning on privileged information $r$ (e.g., a verified reasoning trace).

In OPSD, the teacher and the student are the same model; the teacher is simply granted access to privileged context unavailable to the student, such as a ground-truth solution or terminal feedback from the environment. The standard OPSD objective takes the following dense distillation form:
\begin{equation}
\mathcal{L}_{\text{OPSD}}(\theta) = \mathbb{E}_{y \sim \pi_S} \left[ \frac{1}{|y|} \sum_{t=1}^{|y|} \mathrm{KL}\left( \pi_S(\cdot \mid x, y_{<t}) \,\|\, \mathrm{sg}[\pi_T(\cdot \mid x, r, y_{<t})] \right) \right],
\end{equation}
where $T$ and $S$ denote the same model, and $r$ is the privileged context. Inherited from OPD, the OPSD objective is equivalent to reinforcement learning with per-token advantage $A_{t} = \log \pi_T(y_t \mid x, r, y_{<t}) - \log \pi_S(y_t \mid x, y_{<t})$.

\begin{figure}
    \centering
    \includegraphics[width=0.99\linewidth]{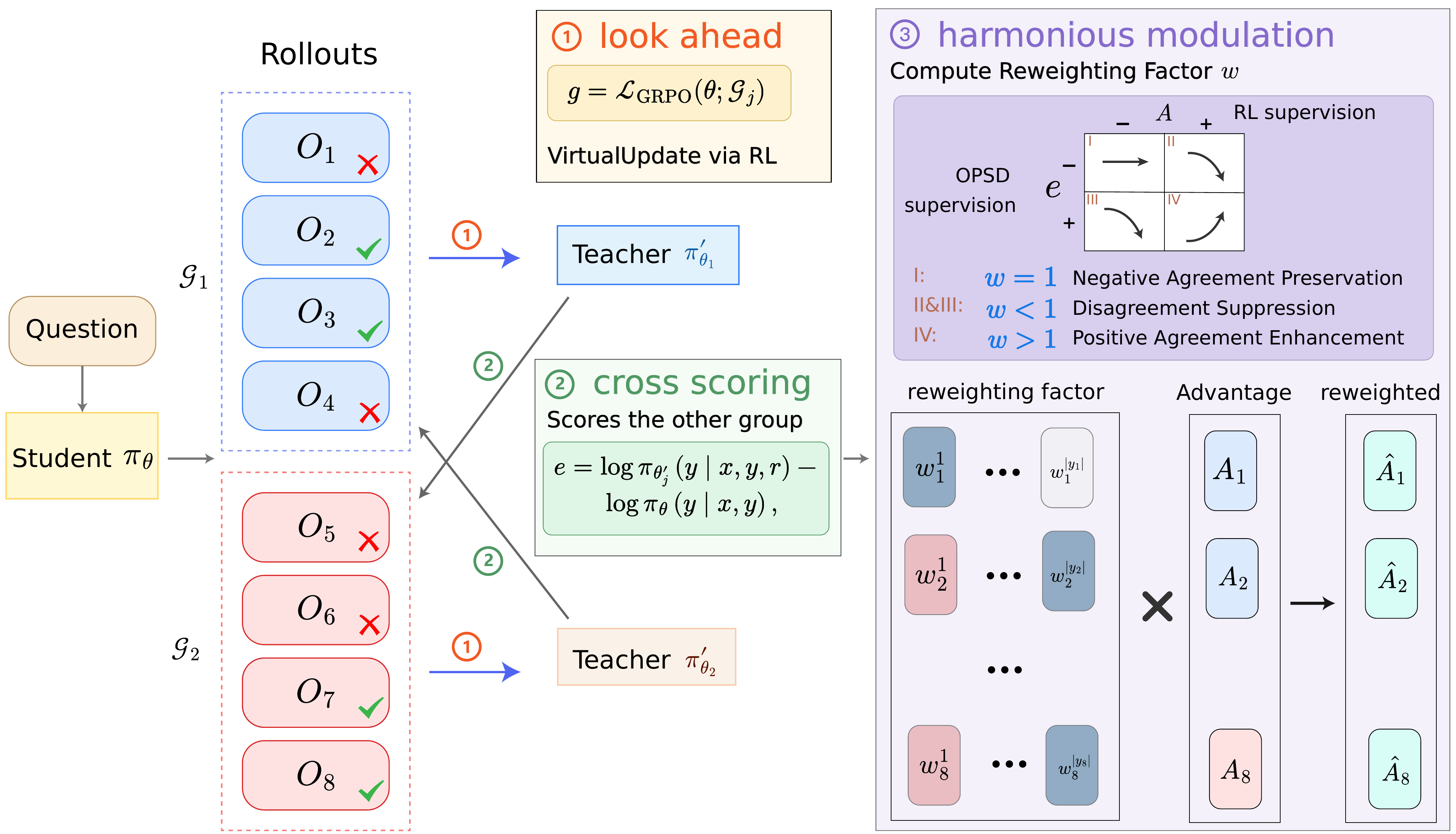}
    \caption{\textbf{Overview of LEAD.} LEAD consists of three steps: Lookahead, Cross Scoring, and Harmonious Modulation. Refer to Section~\ref{sec:method_formulation} for the details.}
    \label{fig:main}
\end{figure}

\section{LEAD: Lookahead Self-Distillation}

\subsection{Formulation}
\label{sec:method_formulation}

The key idea of LEAD is to let the teacher first acquire information from the current on-policy batch through a lookahead update, and then use the resulting lookahead teacher to provide privileged token-level supervision. This differs from conventional OPSD, where the teacher supervision for a rollout is determined only by privileged information associated with that same rollout.

To understand why such a lookahead teacher can provide additional supervision, consider a batch of \(N\) on-policy rollouts
$
\mathcal{B}
=
\left\{
z_i=(x_i,y_i,r_i)
\right\}_{i=1}^{N}.
$
For a token \(y_{i,t}\), we then examine how post update from model parameter $\pi_{\theta}$ to $\pi_{\theta^+}$ with this rollout changes the privileged log-probability of a token from rollout \(i\). Define the post-update privileged log-probability increment
\begin{equation}
\Delta_{i,t}
=
\log \pi_{\theta^+}(y_{i,t}\mid x_i,y_{i,<t},r_i)
-
\log \pi_{\theta}(y_{i,t}\mid x_i,y_{i,<t},r_i).
\end{equation}
\noindent\textbf{Proposition 1} (Post-update Privileged Log-probability Decomposition). For a token \(y_{i,t}\) in rollout \(i\), after a first-order policy-gradient update on the whole batch $\mathcal{B}$, the post-update privileged log-probability increment on the \(y_{i,t}\) admits the decomposition
\begin{equation}
\Delta_{i,t}
=
\underbrace{
\eta A_i {h_{i,t}}^{\top}V_i
}_{\text{self-induced}}
+
\underbrace{
\eta\sum_{j\neq i}
A_j {h_{i,t}}^{\top}V_j
}_{\text{cross-induced}}
+
o(\eta).
\end{equation}
where 
$
u_{i,t}
=
\log \pi_\theta(y_{i,t}\mid x_i,y_{i,<t}),
$ and $
s_{i,t}
=
\log \pi_\theta(y_{i,t}\mid x_i,y_{i,<t},r_i)
$ define the RL log-probability and the privileged log-probability,
with corresponding gradients
$
v_{i,t}=\nabla_\theta u_{i,t},
h_{i,t}=\nabla_\theta s_{i,t}
$ respectively.
$
V_i=\sum_k v_{i,k}
$
denote the aggregate RL gradient of rollout \(i\). \(A_i\in\mathbb{R}\) denotes the advantage of rollout \(i\) and $\eta$ is the learning rate of RL update. Refer to Appendix \ref{app:proof}
for detailed proof.

Proposition 1 reveals that the post-update privileged log-probability increment of an instance consists of two distinct components. The self-induced term reflects how the RL signal of the instance itself changes its privileged prediction, whereas the cross-induced term captures how learning from other instances affects the privileged prediction of the current instance.

Importantly, the self-induced term introduces a phenomenon we refer to as \emph{supervision entanglement}: the RL signal first updates the model based on this instance, and the updated model is then used to evaluate the privileged prediction of the very same instance. Such supervision entanglement can amplify instance-specific biases that are not necessarily beneficial for generalization. In contrast, the cross-induced term arises from the interaction between the privileged prediction of one instance and the RL learning signal from other instances. The resulting supervision therefore reflects information acquired across instances rather than repeatedly reusing supervision from the same instance.

% Importantly, the self-induced term introduces a spuersion recurrent entanglement phenomenon: between the privileged information and the RL signal of the same instance, both of which provide supervision about the same underlying outcome, such as the ground-truth solution or environment feedback. Consequently, the same instance-level supervision is incorporated through two coupled pathways. Such repeated coupling may amplify instance-specific biases that are not necessarily beneficial for generalization. In contrast, the cross-induced term arises from the interaction between the privileged prediction of one instance and the RL learning signal from other instances, thereby introducing supervision acquired across instances.

Motivated by this observation, LEAD introduces a \emph{cross-scoring} design: the teacher is first updated using one set of on-policy rollouts and then used to score another set of rollouts. This design explicitly exploits the inter-instance distillation while avoiding the supervision entanglement, as detailed in the following section.

\subsection{LEAD}

As shown in Figure~\ref{fig:main}, given an on-policy batch of queries and their sampled rollouts, LEAD proceeds through three key steps. First, in \emph{Lookahead}, we split the on-policy rollouts into subsets and perform a virtual RL update on each subset, allowing the model to acquire information from the corresponding rollouts without modifying the actual model parameters. Second, in \emph{Cross Scoring}, the resulting lookahead teacher from one subset is used to evaluate rollouts from another subset, thereby transferring learning signals across instances and explicitly capturing the inter-instance supervision identified in Proposition~1. Finally, in \emph{Harmonious Modulation}, we combine the resulting OPSD signal with the RLVR signal, leveraging their complementary information while accounting for the potential conflict between the two supervision signals.

\subsubsection{Lookahead}
Let \(\theta\) denote the current policy parameters and let
\(\mathcal{L}_{\mathrm{GRPO}}(\theta;\mathcal{G})\) denote the GRPO objective trained on a subset of rollouts \(\mathcal{G}\). 
Concretely, consider an on-policy batch consisting of $Q$ queries with $G$ sampled rollouts each, so that $N=QG$. We partition the rollouts along the rollout dimension into two disjoint groups of $\tfrac{1}{2}QG$ rollouts each, denoted by $\mathcal{G}_1$ and $\mathcal{G}_2$.

For each group \(\mathcal{G}_j\), we first use the RL gradient to construct a \emph{lookahead} parameter state that approximates the model after learning from \(\mathcal{G}_j\):
\begin{equation}
\theta'_j
=
\operatorname{VirtualUpdate}(\theta, \mathcal{L}_{\mathrm{GRPO}}(\theta;\mathcal{G}_j)), \qquad j\in\{1,2\}.
\end{equation}
In practice, \(\operatorname{VirtualUpdate}\) follows the same parameter-update rule as the underlying optimizer, including its accumulated optimization statistics, while keeping the original model parameters and the optimizer state unchanged. Thus, \(\theta'_j\) represents a learned model that has taken one RL update on the rollouts in \(\mathcal{G}_j\).

\subsubsection{Cross Scoring}
After constructing the two lookahead teachers, we evaluate each group using the teacher obtained from the other group. Specifically, for a rollout
$
y_i
$
belonging to \(\mathcal{G}_2\), its token-level OPSD supervision under the lookahead teacher \(\theta'_1\) is
\begin{equation}
e_{i,t}
=
\log \pi_{\theta'_1}\left(y_{i,t}\mid x_i,y_{i,<t},r_i\right)
-
\log \pi_{\theta}\left(y_{i,t}\mid x_i,y_{i,<t}\right),
\end{equation}
Similarly, rollouts in \(\mathcal{G}_1\) are scored by \(\theta'_2\). 
Since the magnitude of \(e_{i,t}\) depends on the scale of the virtual update, we normalize the signal within the batch:
\begin{equation}
\hat{e}_{i,t}
=
\operatorname{clip}
\left(
\frac{
e_{i,t}-\mu_e
}{
\sigma_e+10^{-6}
},
-c,c
\right),
\end{equation}
where \(\mu_e\) and \(\sigma_e\) denote the batch mean and standard deviation of $e_{i,t}$, and \(c\) is a hyperparameter that controls the maximum magnitude of the normalized signal. This normalization removes the dependence of the signal magnitude on the absolute scale of the virtual update while retaining its directional information.

\subsubsection{Harmonious Modulation}
Existing approaches typically treat the two supervision signals independently, overlooking that they may provide complementary yet potentially conflicting guidance during optimization. GRPO provides outcome-based credit assignment from the current on-policy rollouts, whereas LEAD captures how the model's token-level log-probabilities would change after learning from other rollouts. Their learning signals may therefore point in different directions. To reconcile these two sources of supervision, LEAD introduces \emph{Harmonious Modulation}, which adaptively modulates the GRPO advantage according to the agreement between the two signals.

The GRPO advantage is modulated by a reweighting factor:
\begin{equation}
\hat A_{i,t}=w_{i,t}A_{i,t} .
\end{equation}
\emph{Positive Agreement Enhancement} applies when both the lookahead signal and the GRPO advantage are positive. Since the two signals reinforce the same positive learning direction, we increase the magnitude of the advantage accordingly. \emph{Disagreement Suppression} applies when the two signals point in opposite directions. We reduce the magnitude of the GRPO advantage to mitigate the conflict while preserving the outcome-based RL direction. Finally, \emph{Negative Agreement Preservation} applies when both signals are non-positive. In this case, we leave the advantage unchanged rather than further amplifying a negative update. Specifically, we define the reweighting factor as
\begin{equation}
\label{eq:gate}
w_{i,t}=
\begin{cases}
1+\rho_a\dfrac{|\hat e_{i,t}|}{c},
& \hat e_{i,t}>0,\ A_{i,t}>0, \quad \text{(i) Positive Agreement Enhancement}\\[6pt]
1-\rho_b\dfrac{|\hat e_{i,t}|}{c},
& \hat e_{i,t} A_{i,t}<0, \quad \text{(ii) Disagreement Suppression}\\[6pt]
1,
& \text{otherwise.} \quad\qquad\ \ \text{(iii) Negative Agreement Preservation}
\end{cases}
\end{equation}

where $\rho_a\geq0$ and $\rho_b\in[0,1]$ control the strengths of positive-agreement enhancement and disagreement suppression, respectively. Since $\hat e_{i,t}$ is clipped to $[-c,c]$, the normalized magnitude $|\hat e_{i,t}|/c$ lies in $[0,1]$. 

Therefore, the reweighting factor satisfies
\begin{equation}
w_{i,t}\in[1-\rho_b,\,1+\rho_a].
\end{equation}
where $\rho_a$ determines the maximum relative amplification under positive agreement, whereas $\rho_b$ determines the maximum relative attenuation under disagreement. 

LEAD only modulates the magnitude of the GRPO advantage without changing its sign. The resulting $\hat A_{i,t}$ is then used in place of $A_i$ in the standard GRPO objective.

\section{Experiments}
\label{sec:exp}

\subsection{Setups}
\label{sec:exp_setup}

\paragraph{Models and Benchmarks.}
We evaluate all methods with two Qwen3~\citep{yang2025qwen3} backbone models of different scales, Qwen3-4B and Qwen3-8B, to examine whether the benefits of LEAD persist across model sizes. We evaluate LEAD on two complementary datasets: \emph{scientific reasoning}, which focuses on single-response reasoning with outcome-based rewards, and \emph{interactive agents}, which requires long-horizon, multi-turn decision making with sparse and delayed rewards. Together, these benchmarks allow us to examine whether the benefits of inter-instance supervision extend from conventional reasoning tasks to more challenging agentic settings.
\begin{itemize}[leftmargin=*, itemsep=2pt, topsep=2pt, parsep=0pt]
  \item \textbf{Scientific reasoning.} Following prior works~\citep{hubotter2026reinforcement,yang2026self}, we train on four datasets from the reasoning subset (L3) of SciKnowEval~\citep{feng2024sciknoweval}, covering undergraduate-level Biology, Chemistry, Materials Science, and Physics. 
  
  \item  \textbf{Interactive agents.} We further evaluate on three agent benchmarks following SDAR~\citep{lu2026sdar}: ALFWorld~\citep{shridhar2021alfworld} for embodied household tasks, WebShop~\citep{yao2022webshop} for web navigation and purchase, and Search~\citep{jin2025searchr1} for tool-augmented question answering. This setting provides a more challenging test of LEAD because each training example consists of multiple agent turns with sparse, end-of-episode rewards. 
\end{itemize}

\paragraph{Scientific Reasoning Dataset Curation.}
SciKnowEval~\citep{feng2024sciknoweval} is widely used by prior works~\citep{hubotter2026reinforcement,yang2026self} for evaluations. However, we identified inconsistencies in the original dataset release, including malformed options and questions without a valid answer. We therefore perform a systematic audit using independent judgments from GPT-5.5 and Claude Opus 5. For the training split, items with consistent judgments are retained, relabelled, or minimally rewritten, while inconsistent cases are removed. The curation primarily affects Physics: approximately $27\%$ of its training items require correction or removal, compared with less than $4\%$ in Biology, Chemistry, and Materials Science combined. The test split is affected far less: 4.8\% of its items are removed and  2.7\% are corrected. Human verification further shows $99.3\%$ agreement with the curated answers across randomly sampled items. Detailed curation criteria, statistics, and representative cases are provided in Appendix~\ref{app:data}.

% \paragraph{Evaluation Settings.}
% Following \cite{pan2026rlcsd}, we adopt a unified evaluation protocol to separately assess performance on the training domain and generalization to unseen domains. For each method and benchmark, we train one model per domain from the corresponding base checkpoint and evaluate the resulting model on all domains within the same suite. This allows each training run to provide both in-domain and out-of-domain measurements:
% \begin{itemize}[leftmargin=*, itemsep=2pt, topsep=2pt, parsep=0pt]
% \item \textbf{In-domain evaluation.} We evaluate the model on the same domain used for training, measuring its performance on the target domain.

% \item \textbf{Out-of-domain (OOD) evaluation.} We evaluate the same model on the remaining domains in the suite without further training, measuring its cross-domain generalization ability.
% \end{itemize}

\paragraph{Evaluation Settings.}
Following \cite{pan2026rlcsd}, we adopt a unified evaluation protocol to separately assess in-domain performance and cross-domain generalization. For each method and each domain, we train one model using the training split of that domain, starting from the corresponding base model. The resulting model is then evaluated on the test splits of all domains within the same suite. Specifically:

\begin{itemize}[leftmargin=*, itemsep=2pt, topsep=2pt, parsep=0pt]
\item \textbf{In-domain evaluation.} The model is evaluated on the test split of the same domain used for training. This measures performance on the domain seen during training.

\item \textbf{Out-of-domain (OOD) evaluation.} The same model is evaluated on the test splits of all other domains in the suite, without any further training or adaptation. We report the average performance across these unseen domains to measure cross-domain generalization.
\end{itemize}

\paragraph{Metrics.}
On the scientific-reasoning suite we sample $16$ responses per test question and report the mean accuracy, denoted mean@16; a response counts as correct only if the verifier matches it against the ground-truth answer. On the interactive-agent suite, ALFWorld and WebShop are scored by the environment's episode success rate over a fixed held-out set, and Search by the mean exact-match score over its seven constituent QA datasets. All numbers are reported in percent.

Following~\citet{hubotter2026reinforcement}, in-domain scores are the maximum value reached over the evaluation checkpoints of a run, whereas OOD scores are averaged over the last three checkpoints: the former measures how far a method can push the training domain, the latter describes the model obtained at the end of training and is therefore insensitive to a transient peak. Evaluation is run every $10$ global steps on the scientific suite and every $5$ steps on the agent suite.

\paragraph{Baselines.}
We compare LEAD against three baselines: (1) \textbf{GRPO}~\citep{shao2024deepseekmath} group relative policy optimization with binary outcome rewards verified against ground-truth answers; (2) \textbf{SDPO}~\citep{hubotter2026reinforcement} is a representative \emph{dense} OPSD method, which constructs dense-logit distillation using Jensen–Shannon divergence;  (3) \textbf{RLSD}~\citep{yang2026self} that uses the teacher–student probability gap on sampled tokens to modulate advantage.

\paragraph{Implementation Details.}
For each dataset in scientific-reasoning suite, we sample $G=8$ on-policy rollouts with a training batch of $32$ queries ($256$ rollouts per global batch). For the agent suite, we follow the protocol of SDAR~\citep{lu2026sdar} and refer to Appendix~\ref{app:implementation} for detailed implementation.
% These methods share the same privileged context in each suite: a correct rollout from the same group on the scientific suite, and, following~\citet{lu2026sdar}, a natural-language skill document for the task type on the agent suite, which describes \emph{how} to act rather than revealing the answer. We train for $10$ epochs per domain using AdamW.
For LEAD, the virtual RL update uses an inner learning rate of $\eta_{\text{in}}=10^{-5}$ to obtain the lookahead teacher parameters. We disable this virtual update for the first $10$ global steps and use a first-moment estimate consisting of $0.7$ historical momentum and $0.3$ of the current half-batch gradient. The resulting virtual RL update is applied only to construct the lookahead teacher and does not modify the actual model parameters. The lookahead signal is clipped at $c=3.0$, with $\rho_a=0.1$ and $\rho_b=0.3$; $\rho_b$ is linearly annealed to zero over the first $70\%$ of training. More implementation details are provided in Appendix~\ref{app:implementation}.

\begin{table*}[t]
\centering
\small
\caption{\textbf{In-domain results across the scientific-reasoning and interactive-agent suites.} Each result is evaluated on the test split of the same domain used for training. Maximum mean@16 (\%) and success rate (\%) is reported on the scientific-reasoning and interactive-agent suites, respectively. \emph{Base} is the untrained model and is listed for reference and excluded from the ranking. \textbf{Bold} indicates the best result among the four training methods and \underline{underline} the second best. On the scientific-reasoning suite, both the training and the test splits are curated as described in Appendix~\ref{app:data}, and all methods use the same curated splits.}
\label{tab:indomain_all}
\resizebox{\textwidth}{!}{%
\begin{tabular}{c|ccccc|cccc}
\toprule
& \multicolumn{5}{c|}{\textbf{Scientific Reasoning}} &
\multicolumn{4}{c}{\textbf{Interactive Agent}} \\
Method & Bio. & Chem. & Mate. & Phy. & Avg.
& AlfWorld & WebShop & Search & Avg. \\
\midrule
\multicolumn{10}{l}{\texttt{\textbf{Qwen3-4B}}} \\
Base Model        & 31.86 & 43.67 & 61.74 & 79.67 & 54.24
            & 30.68 & 2.73 & 27.03 & 20.15 \\
GRPO        & 52.50 & 78.05 & \underline{80.51} & \textbf{94.77} & 76.46
            & \underline{71.40} & 75.80 & \underline{41.60} & \underline{62.93} \\
SDPO        & 52.12 & 77.48 & 73.72 & 86.99 & 72.58
            & 57.00 & \underline{79.20} & 39.56 & 58.59 \\
RLSD        & \underline{52.88} & \underline{78.51} & 80.38 & \underline{94.67} & \underline{76.61}
            & 68.80 & 78.10 & 40.50 & 62.47 \\
LEAD (ours) & \textbf{53.87} & \textbf{81.19} & \textbf{85.82} & 93.65 & \textbf{78.63}
            & \textbf{72.70} & \textbf{79.30} & \textbf{42.50} & \textbf{64.83} \\
\midrule
\multicolumn{10}{l}{\texttt{\textbf{Qwen3-8B}}} \\
Base Model        & 30.47 & 41.87 & 58.78 & 78.84 & 52.49
            & 18.20 & 5.47 & 18.26 & 13.98 \\
GRPO        & 54.87 & \underline{82.57} & \underline{78.83} & 91.80 & 77.02
            & 71.10 & 84.40 & \underline{41.99} & 65.83 \\
SDPO        & \textbf{60.88} & 81.40 & 77.15 & 87.91 & 76.84
            & \underline{75.00} & 84.40 & 41.19 & 66.86 \\
RLSD        & 56.88 & 81.52 & 78.23 & \underline{92.32} & \underline{77.24}
            & 73.30 & \textbf{88.30} & 41.56 & \underline{67.72} \\
LEAD (ours) & \underline{58.50} & \textbf{82.60} & \textbf{80.70} & \textbf{94.26} & \textbf{79.02}
            & \textbf{84.40} & \underline{86.70} & \textbf{42.19} & \textbf{71.10} \\
\bottomrule
\end{tabular}
}
\end{table*}

% ---------------------------------------------------------------- Table: merged OOD (sci + agent)
\begin{table*}[t]
\centering
\small
\caption{\textbf{Out-of-domain results across the scientific-reasoning and interactive-agent suites.} Each result is the averaged performance of the model trained on the corresponding domain while evaluated on the \emph{remaining} domains of the same suite. \textbf{Bold} indicates the best result among the four training methods and \underline{underline} the second best.}
\label{tab:ood_all}
\resizebox{\textwidth}{!}{%
\begin{tabular}{c|ccccc|cccc}
\toprule
& \multicolumn{5}{c|}{\textbf{Scientific Reasoning}} &
\multicolumn{4}{c}{\textbf{Interactive Agent}} \\
Method & Bio. & Chem. & Mate. & Phy. & Avg.
& AlfWorld & WebShop & Search & Avg. \\
\midrule
\multicolumn{10}{l}{\texttt{\textbf{Qwen3-4B}}} \\
Base Model        & 61.69 & 57.76 & 51.73 & 45.76 & 54.24
            & 14.88 & 28.85 & 16.70 & 20.14 \\
GRPO        & \textbf{60.52} & \textbf{58.89} & \underline{50.06} & \textbf{48.37} & \textbf{54.46}
            & 14.28 & \textbf{30.67} & 20.45 & \textbf{21.80} \\
SDPO        & 33.98 & 33.53 & 25.92 & 38.23 & 32.92
            & 12.93 & 25.05 & \underline{21.35} & 19.78 \\
RLSD        & \underline{59.88} & \underline{58.63} & 48.83 & 47.20 & 53.64
            & \underline{14.43} & \underline{28.41} & 18.02 & 20.29 \\
LEAD (ours) & 58.98 & 57.31 & \textbf{52.72} & \underline{48.07} & \underline{54.27}
            & \textbf{15.01} & 28.09 & \textbf{21.73} & \underline{21.61} \\
\midrule
\multicolumn{10}{l}{\texttt{\textbf{Qwen3-8B}}} \\
Base Model        & 59.83 & 56.03 & 50.39 & 43.71 & 52.49
            & 11.87 & 18.23 & 11.84 & 13.98 \\
GRPO        & \textbf{61.53} & \textbf{59.66} & 50.57 & \underline{45.45} & 54.30
            & 8.65 & 13.19 & \textbf{19.50} & \underline{13.78} \\
SDPO        & 41.72 & 31.35 & 40.26 & 43.41 & 39.19
            & \textbf{11.71} & 11.14 & 11.15 & 11.33 \\
RLSD        & 61.12 & 59.55 & \textbf{53.01} & 45.37 & \underline{54.76}
            & \underline{11.16} & \underline{13.76} & 16.22 & 13.71 \\
LEAD (ours) & \underline{61.21} & \underline{59.64} & \underline{52.79} & \textbf{45.93} & \textbf{54.89}
            & 10.01 & \textbf{14.05} & \underline{18.40} & \textbf{14.15} \\
\bottomrule
\end{tabular}
}
\end{table*}

\subsection{Main Results}
\label{sec:exp_result}

\paragraph{In-Domain Results.}
% All four methods improve substantially over the untrained backbone, which reaches only $54.23$ and $52.49$ on the scientific average and $20.14$ and $13.98$ on the agent average for Qwen3-4B and Qwen3-8B, so the comparison below is between methods that all work. 
As shown in Table~\ref{tab:indomain_all}, LEAD consistently improves in-domain performance across both model scales and task suites. In scientific reasoning, LEAD achieves the highest performance for both Qwen3-4B and Qwen3-8B, outperforming the strongest baseline by $2.02$ and $1.78$ points, respectively. The same trend extends to interactive agents. On Qwen3-4B, LEAD is the best method on all three benchmarks ($72.70$, $79.30$ and $42.50$), giving an agent average of $64.83$ against $62.93$ for GRPO and $62.47$ for RLSD. On Qwen3-8B, LEAD is the best on ALFWorld ($84.40$) and Search ($42.19$) and the second best on WebShop. These results suggest that the inter-instance supervision captured by the lookahead teacher remains beneficial across both conventional reasoning and sparse-reward, multi-turn agent tasks.

\paragraph{Out-of-Domain Results.}
Table~\ref{tab:ood_all} shows that LEAD largely preserves the OOD performance of GRPO while achieving substantially higher in-domain accuracy. On scientific reasoning, LEAD obtains OOD averages of $54.27$ and $54.89$ on Qwen3-4B and Qwen3-8B, nearly matching GRPO at the smaller scale and achieving the best OOD performance at the larger scale, while improving its in-domain averages over GRPO by $2.17$ and $2.00$ points as shown in Table~\ref{tab:indomain_all}. In contrast, SDPO exhibits a substantial OOD degradation at both scales, whereas RLSD remains close to GRPO, suggesting that directly injecting dense privileged supervision can be detrimental to cross-domain generalization, while bounded advantage modulation better preserves the outcome-supervised learning behavior.
% The \emph{Base Model} rows make this failure mode concrete: SDPO ends $21.32$ and $13.30$ points \emph{below} the untrained backbone on the scientific OOD average, so training does not merely fail to help transfer but actively removes it, while LEAD finishes at or above the base level at both scales ($54.27$ against $54.23$, and $54.89$ against $52.49$). 
% On the interactive-agent suite the cross-benchmark picture is different: all four methods fall to $19$--$22\%$ on Qwen3-4B and $11$--$15\%$ on Qwen3-8B, far below their in-domain scores, and the spread between methods is only a few points: LEAD is second best on Qwen3-4B ($21.46$ against $21.64$ for GRPO) and best on Qwen3-8B ($14.15$). AlfWorld, WebShop and Search differ in action space and observation format far more than the four scientific domains differ from one another, so a policy specialised on one of them transfers little to the others regardless of the training objective. Overall, these results show that LEAD improves in-domain performance without sacrificing the cross-domain generalization of the underlying GRPO training on the scientific suite, while cross-benchmark agent transfer remains an open problem for all methods considered.
On the interactive-agent suite, however, the OOD performance of all methods falls to $19$--$22\%$ on Qwen3-4B and $11$--$15\%$ on Qwen3-8B, far below their in-domain scores. LEAD ranks second on Qwen3-4B ($21.61$ vs. $21.80$ for GRPO) and achieves the highest OOD performance on Qwen3-8B ($14.15$). In contrast to the scientific domains, ALFWorld, WebShop, and Search differ substantially in their action spaces and observation formats, making cross-benchmark transfer more challenging and limiting the benefit of any single training objective. Overall, these results show that LEAD improves in-domain performance without sacrificing the generalization.

\subsection{Training Dynamics}
\label{sec:dynamics}

Figures~\ref{fig:dyn_reward} and~\ref{fig:dyn_eval} provide a step-by-step view of training on Chemistry and Materials Science, complementing the final results reported in Tables~\ref{tab:indomain_all} and~\ref{tab:ood_all}. LEAD consistently achieves strong training rewards and in-domain performance throughout training, with faster convergence in the early stages and a sustained advantage toward the end of training. Meanwhile, its out-of-domain performance remains relatively stable across the training process, in contrast to the progressive degradation observed for SDPO. Overall, these dynamics show that LEAD accelerates in-domain learning while better preserving performance on unseen domains throughout training.

\subsection{More Analysis}
\label{sec:ablation}

We conduct the ablation study on the scientific-reasoning suite with Qwen3-4B, keeping all other hyperparameters identical to the main experiments. Table~\ref{tab:ablation} evaluates the contribution of the three key components of LEAD: \emph{Lookahead}, \emph{Cross-Scoring}, and \emph{Harmonious Modulation}.

\paragraph{Lookahead.}
Removing the lookahead update (\emph{w/o lookahead}) reduces the in-domain average from $78.63$ to $76.10$, a drop of $2.53$ points. This variant uses the current model $\theta$ directly as the teacher, retaining the privileged-context supervision without allowing the teacher to first acquire information from the on-policy batch. The degradation across Biology, Chemistry, and Materials Science indicates that the additional information acquired through the virtual RL update contributes meaningfully to the learning signal. The OOD average also decreases slightly from $54.27$ to $54.09$.

\paragraph{Cross-Scoring.}
To isolate the effect of cross-scoring, we replace the cross-group scoring scheme with a self-induced variant (\emph{w/ self-induced}), where each lookahead teacher scores the same rollouts that were used to construct it. This modification reintroduces the self-induced component identified in Proposition~1 while preserving the lookahead update itself. The in-domain average drops from $78.63$ to $76.90$, corresponding to a $1.73$-point degradation. This result highlights the importance of using information acquired from other instances rather than repeatedly coupling the privileged signal with the RL signal from the same instance. 

\paragraph{Harmonious Modulation.}
We further ablate the three cases defined in Eq.~\ref{eq:gate}: \emph{Positive Agreement Enhancement} (i), \emph{Disagreement Suppression} (ii), and \emph{Negative Agreement Preservation} (iii). Removing \emph{Positive Agreement Enhancement} (\emph{w/o (i)}) reduces the in-domain average to $77.61$ and the OOD average to $52.06$; this variant reaches the highest Biology ($57.12$) and Physics ($93.85$) scores of the table, but its average over the four domains remains below LEAD's. Removing \emph{Disagreement Suppression} for positive advantages (\emph{w/o (ii) ($A>0$)}) reduces the in-domain average to $75.90$, while the OOD average is $54.03$. The corresponding ablation for negative advantages (\emph{w/o (ii) ($A<0$)}) reduces the in-domain average to $77.05$, while the OOD average is $53.50$. Finally, removing \emph{Negative Agreement Preservation} (\emph{w/o (iii)}) reduces the in-domain average to $76.16$ and the OOD average to $53.83$. These results suggest that the effectiveness of Harmonious Modulation comes from coordinating the three cases: strengthening positive agreement, suppressing disagreement, and preserving the magnitude of negative updates when the two signals agree.

\begin{table*}[t]
\centering
\small
\caption{\textbf{Ablation study on the scientific-reasoning suite (Qwen3-4B).} In-domain is the maximum mean@16 (\%) reached during training on that domain; out-of-domain is the accuracy of the same model on the remaining three domains, averaged over the last three evaluation checkpoints. \textbf{Bold} indicates the best result in each column and \underline{underline} the second best.}
\label{tab:ablation}
\resizebox{\textwidth}{!}{%
\begin{tabular}{l|ccccc|ccccc}
\toprule
& \multicolumn{5}{c|}{\textbf{In-domain}} & \multicolumn{5}{c}{\textbf{Out-of-domain}} \\
\cmidrule(lr){2-6}\cmidrule(lr){7-11}
Variant & Bio. & Chem. & Mate. & Phy. & Avg. & Bio. & Chem. & Mate. & Phy. & Avg. \\
\midrule
LEAD               & \underline{53.87} & \textbf{81.19} & \textbf{85.82} & \underline{93.65} & \textbf{78.63}
                            & 58.98 & 57.31 & \underline{52.72} & 48.07 & \textbf{54.27} \\
\midrule
\multicolumn{11}{l}{\emph{Lookahead}} \\
\quad w/o lookahead       & 51.50 & 79.34 & 80.51 & 93.06 & 76.10
                            & \textbf{60.04} & 56.23 & 51.60 & \underline{48.49} & \underline{54.09} \\
\midrule
\multicolumn{11}{l}{\emph{Cross-Scoring}} \\
\quad w/ self-induced   & 52.38 & \underline{80.92} & 81.45 & 92.83 & 76.90
                            & 59.25 & 56.05 & \textbf{53.06} & 46.43 & 53.70 \\
\midrule
\multicolumn{11}{l}{\emph{Harmonious Modulation}} \\
\quad w/o (i)           & \textbf{57.12} & 78.77 & 80.71 & \textbf{93.85} & \underline{77.61}
                            & 55.00 & 57.27 & 50.57 & 45.41 & 52.06 \\
\quad w/o (ii) ($A>0$) & 51.12 & 79.16 & 80.71 & 92.62 & 75.90
                            & 59.17 & \textbf{57.84} & 51.19 & 47.93 & 54.03 \\
\quad w/o (ii) ($A<0$) & 51.75 & 80.38 & \underline{82.53} & 93.55 & 77.05
                              & 56.56 & \underline{57.58} & 51.20 & \textbf{48.64} & 53.50 \\
\quad w/o (iii)     & 49.75 & 80.47 & 81.38 & 93.03 & 76.16
                            & \underline{59.67} & 56.82 & 51.45 & 47.38 & 53.83 \\
\bottomrule
\end{tabular}
}
\end{table*}

\paragraph{Computational Cost.}
LEAD introduces an additional lookahead update compared with prior OPSD methods. Table~\ref{tab:cost} reports the wall-clock time per training step on the scientific-reasoning suite with Qwen3-4B. The additional overhead of LEAD is small: it is $7.8\%$ slower than GRPO and $6.0\%$ slower than RLSD. This overhead is expected to be even smaller for long-reason tasks, such as mathematics and code generation, where rollout generation accounts for a larger fraction of the total step time. In comparison, SDPO is $16.5\%$ slower than GRPO and $8.0\%$ slower than LEAD. This larger overhead likely stems from its dense token-level distillation over the top-$k$ vocabulary, which makes the optimization stage more expensive than the sampled-token-level modulation used by RLSD and LEAD.

\begin{table}[t]
\centering
\small
\caption{\textbf{Wall-clock time per training step (seconds), scientific-reasoning suite with Qwen3-4B}, averaged over the four domains. \emph{Generation} is the rollout stage; \emph{Update} covers the teacher forward passes, the log-probability recomputation and the policy update.}
\label{tab:cost}
\begin{tabular}{lccc}
\toprule
Method & Generation (s/step) & Update (s/step) & Total step (s/step) \\
\midrule
GRPO        & 16.30 & 9.06  & {25.36} \\
SDPO        & 16.03 & 13.52 & 29.55 \\
RLSD        & 15.88 & 9.92  & {25.80} \\
LEAD (ours) & 15.78 & 11.57 & 27.35 \\
\bottomrule
\end{tabular}
\end{table}

\section{Conclusion} 
We introduced LEAD, a novel on-policy self-distillation method that enables the teacher to first learn from other on-policy rollouts before teaching the student. Results on reasoning and agent tasks demonstrate consistent in-domain improvements and strong OOD robustness.

\section*{AI use statement}
In this work, we used generative AI tools for aiding and polishing the writing of the manuscript, including improving grammar, clarity, readability, and presentation.
We have not used generative AI tools for research design, data collection, data analysis, experimental execution, or the generation or verification of experimental results, and the rest of the required disclosure tasks are not applicable to this work.
We have reviewed all AI-assisted work and verified the accuracy and consistency of the resulting text. We take responsibility for the final content of this work, including text, claims, and artifacts produced with the aid of generative AI.
 
\bibliography{iclr2027_conference}
\bibliographystyle{iclr2027_conference}

\appendix

\begin{figure}[t]
\centering
\includegraphics[width=\linewidth]{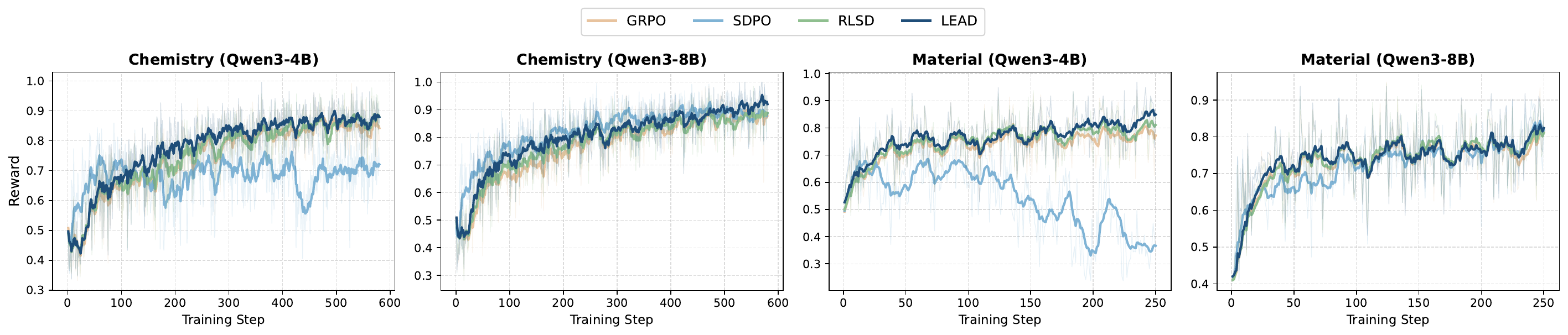}
\caption{Reward curve during training on Chemistry and Materials Science datasets of the scientific-reasoning suite.}
\label{fig:dyn_reward}
\end{figure}

\begin{figure}[t]
\centering
\includegraphics[width=\linewidth]{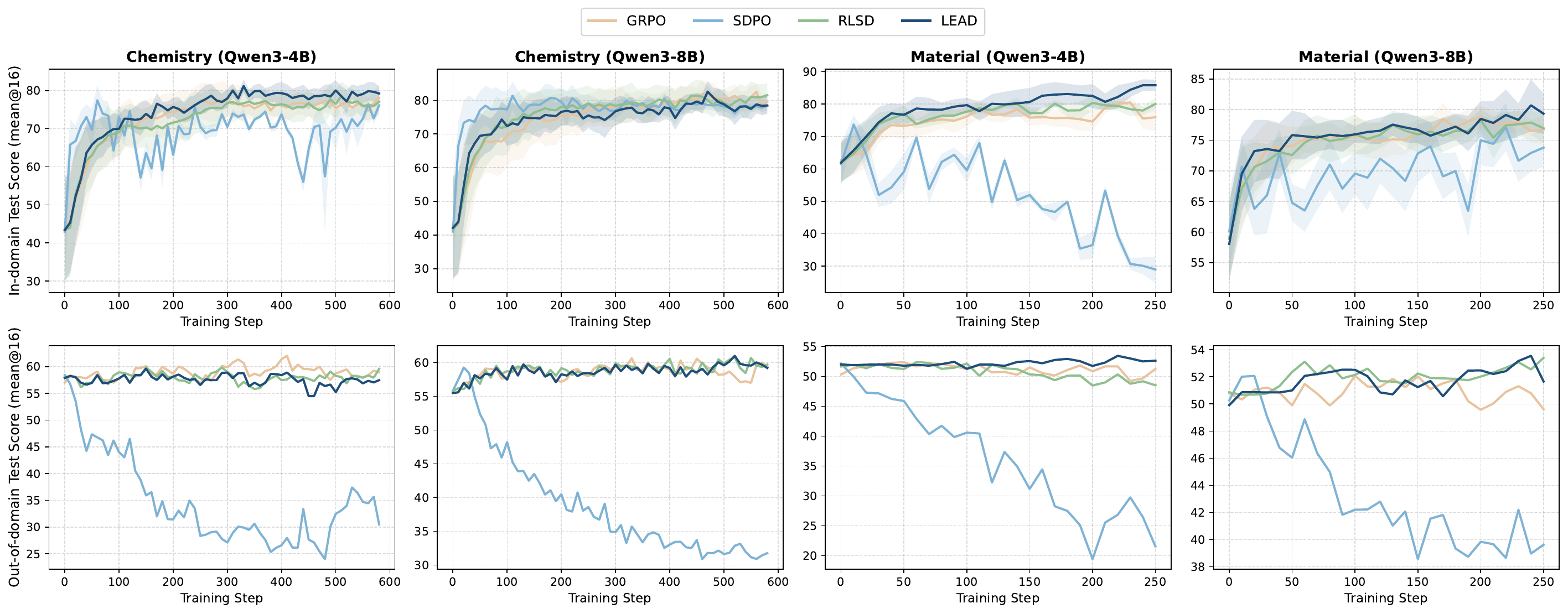}
\caption{Evaluation dynamics on the scientific-reasoning suite. Top row: in-domain mean@16 on the training domain. Bottom row: out-of-domain mean@16, averaged over the three domains that were not used for training.}
\label{fig:dyn_eval}
\end{figure}

\section{Related Work}
\label{app:related}

\subsection{RLVR and On-Policy Distillation} 
Reinforcement learning with verifiable rewards (RLVR)~\citep{shao2024deepseekmath,yu2026dapo,liu2025understanding,zheng2025group,pan2025d} has become an important post-training paradigm for large language models, particularly for reasoning. GRPO~\citep{shao2024deepseekmath} estimates advantages by normalizing rewards across multiple rollouts sampled from the same query. However, the resulting supervision remains at the response level: a single scalar advantage is shared by all tokens, providing limited credit assignment for intermediate reasoning steps.
On-policy distillation (OPD)~\citep{gu2024minillm,agarwal2024policy} provides a complementary token-level learning signal by measuring the distributional gap between a teacher and the student along the student's own trajectories. Most existing OPD methods use reverse-KL, while subsequent work explores alternative divergences, black-box teachers, and more efficient distillation procedures~\citep{ye2025black,wu2026lightning,ko2024distillm}. Recent work has further combined token-level OPD signals with outcome-based RLVR supervision~\citep{xiao2026mimo}.

Our method builds on both RLVR and on-policy distillation. It retains the outcome-based optimization direction of RLVR while introducing token-level supervision through a lookahead self-distillation signal.

\subsection{On-Policy Self-Distillation}
Despite success, OPD introduces considerable computational overhead due to relying on a separate teacher model, which is typically substantially larger than the student. In addition, OPD performs token-level distribution matching over a shared vocabulary, requiring the teacher and student to adopt the same vocabulary spaces. To address these limitations, on-policy self-distillation (OPSD) has emerged~\citep{zhao2026self,han2026adaptive,yu2026preference,hao2026self}. In OPSD, the teacher and the student are the same model; the teacher is simply granted access to privileged context unavailable to the student, such as a ground-truth solution or terminal feedback from the environment. 
Most existing OPSD methods perform dense per-token distillation, using either forward KL or Jensen--Shannon divergence to match the teacher and student distributions~\citep{zhao2026self,hubotter2026reinforcement}. SRPO~\citep{li2026unifying} further introduces sample-level routing, applying GRPO to correct samples and dense distillation to incorrect ones.

More recently, RLSD~\citep{yang2026self,pan2026rlcsd} moves away from dense distribution matching and instead incorporate OPSD with RLVR. RLSD argues that OPSD should complement rather than replace the verifier-derived learning signal. Specifically, the RLVR determines the update direction, while the OPSD signal only modulates its magnitude. This design preserves the outcome-grounded direction of RLVR while incorporating finer-grained privileged supervision.

Nevertheless, these approaches still follow the previous OPSD paradigm, where the supervision signal is derived solely from the privileged information of the same instance. LEAD instead obtains supervision from other rollouts through lookahead updates and cross-scoring, extending on-policy self-distillation from the intra-instance to the inter-instance setting.

\begin{table}[t]
\centering
\small
\caption{\textbf{Curation results for the SciKnowEval L3 reasoning subset.}
\emph{Unchanged} denotes items whose original question, options, and ground-truth label are retained.
\emph{Relabelled} denotes items for which the ground-truth letter is corrected while the question and options remain unchanged.
\emph{Rewritten} denotes items for which both models agree that none of the provided options is correct and derive the same answer; the labelled option is then replaced with the derived answer.
\emph{Removed} denotes items for which the two models reach inconsistent judgments.}
\label{tab:curation}
\resizebox{0.9\textwidth}{!}{%
\begin{tabular}{l|ccccc|ccccc}
\toprule
& \multicolumn{5}{c|}{Training split} & \multicolumn{5}{c}{Test split} \\
Domain & Original & Unchanged & Relabelled & Rewritten & Removed
       & Original & Unchanged & Relabelled & Rewritten & Removed \\
\midrule
Biology   & $450$     & $447$     & $0$  & $0$  & $3$   & $50$  & $50$  & $0$ & $0$ & $0$ \\
Chemistry & $1{,}890$ & $1{,}856$ & $1$  & $4$  & $29$  & $210$ & $208$ & $1$ & $0$ & $1$ \\
Material  & $841$     & $801$     & $1$  & $0$  & $39$  & $94$  & $93$  & $0$ & $0$ & $1$ \\
Physics   & $720$     & $524$     & $26$ & $64$ & $106$ & $80$  & $50$  & $4$ & $7$ & $19$ \\
\midrule
Total     & $3{,}901$ & $3{,}628$ & $28$ & $68$ & $177$ & $434$ & $401$ & $5$ & $7$ & $21$ \\
\bottomrule
\end{tabular}}
\end{table}

\section{Dataset Curation}
\label{app:data}

The reasoning subset (L3) of SciKnowEval~\citep{feng2024sciknoweval} is released as four-way multiple-choice questions with a single letter as the ground truth, covering undergraduate-level Biology, Chemistry, Materials Science, and Physics. It is widely used by prior works~\citep{hubotter2026reinforcement,yang2026self}. 

\paragraph{Noise in Dataset.}
We identified inconsistencies in the original dataset release. As shown in Table~\ref{app_tab:cases_phys} and Table~\ref{app_tab:cases_other}, some items contain part of the question stem as one of the four answer options, with the malformed option also marked as the answer key. Other items contain no valid answer among the provided options.
To ensure the reliability and consistency of our evaluation, we therefore audited all items across the four domains in both the training and test splits, corrected identifiable issues, and removed the remaining problematic items. 

\paragraph{Data Curation.}
Each item is independently reviewed by GPT-5.5 and Claude Opus 5. The two models determine whether a valid answer exists among the provided options and, when applicable, identify the correct answer. We retain and curate an item only when the two models reach the same conclusion; otherwise, the item is removed. Specifically, we consider the following three cases:

\begin{itemize}[leftmargin=*, itemsep=2pt, topsep=2pt, parsep=0pt]
\item \textbf{A valid answer exists.} If both models identify the same option as correct, we compare it with the original label. If the identified option agrees with the original label, the item is left unchanged. If it differs, we \emph{relabel} the item by updating only the ground-truth letter, while leaving the question stem and all options unchanged.
\item \textbf{No valid answer exists.} If both models conclude that none of the provided options is correct and independently derive the same correct answer, we \emph{rewrite} the labelled option using the derived answer. The question stem, option positions, and ground-truth letter remain unchanged.
\item \textbf{Inconsistent judgment.} If the two models disagree on whether a valid answer exists, identify different correct answers, or derive different answers when no provided option is correct, the item is \emph{removed} rather than further modified.
\end{itemize}

This procedure follows a principle of minimal intervention on both splits: among items with consistent judgments, only the ground-truth label or the text of the labelled option is modified, and the question stem is never altered. 
% The test split is curated under the same three cases but is affected far less: of its $434$ items, $21$ are removed and $12$ are corrected ($5$ relabelled and $7$ rewritten), leaving $413$ items of which $401$ are identical to the original release. 

\paragraph{Curation Results.}
As summarized in Table~\ref{tab:curation}, the curation primarily affects the Physics domain. Among the $273$ training items that are either corrected or removed, $196$ ($71.8\%$) are from Physics, including $90$ relabelled or rewritten items and $106$ removed items. By comparison, Biology requires no corrections and only $3$ items are removed. In Physics, $524$ of the $720$ training items ($72.8\%$) remain unchanged, $26$ are relabelled, $64$ are rewritten, and $106$ are removed, leaving $614$ curated items. The test split is affected far less: $21$  ($4.8\%$) of its $434$ items are removed and $12$  ($2.7\%$) are corrected, leaving $50$, $209$, $93$, and $61$ curated items for Biology, Chemistry, Materials Science, and Physics, respectively, compared with $50$, $210$, $94$, and $80$ items in the original release. All $7$ rewritten test items and $4$ of the $5$ relabelled ones are in Physics. Refer to Table~\ref{app_tab:cases_phys} and Table~\ref{app_tab:cases_other} for detailed cases.

\paragraph{Human Verification.}
To ensure the correctness of curation results, we randomly selected 100 samples from the curated samples from dataset. 9 human experts (with bachelor or higher degree) are divided into 3 groups, each with 3 person. They were asked to judge the following question: Whether the answers to the questions was correct. The human verification with an accuracy rate of 100.0\%, 99.0\% and 99.0\%, with an averaged  accuracy of 99.3\%, indicating the effectiveness of our curation method.

\subsection{Detailed Implementation}
\label{app:implementation}

\begin{table}[h]
\centering
\caption{\textbf{Method-specific hyperparameters.} Training hyperparameters for GRPO, SDPO, RLSD, and LEAD.}
\label{app:method_para}
\resizebox{0.8\textwidth}{!}{%
\begin{tabular}{l|cccc}
\toprule
 & GRPO & SDPO & RLSD & LEAD (ours) \\
\midrule
Learning rate & $1${\scriptsize e}$\text{-}6$ & $1${\scriptsize e}$\text{-}5$ & $1${\scriptsize e}$\text{-}6$ & $1${\scriptsize e}$\text{-}6$ \\
% PPO micro-batch per GPU & $8$ & $2$ & $8$ & $8$ \\
Entropy coefficient & $0$ & $0$ & $0.001$ & $0$ \\
KL loss coefficient & $0.001$ & off & $0.01$ & $0.001$ \\
\midrule
Optimizer & \multicolumn{4}{c}{AdamW} \\
Clip ratio (low, high) & \multicolumn{4}{c}{$0.2$, $0.28$; dual-clip $c_{\text{clip}}=3.0$} \\
Sampling parameters (training) & \multicolumn{4}{c}{temperature $1.0$ with $16$ samples at top-$p$ $0.95$} \\
Sampling parameters (evaluation) & \multicolumn{4}{c}{temperature $0.6$ with $16$ samples at top-$p$ $0.95$} \\
Privileged context (scientific reasoning) & \multicolumn{4}{c}{a correct rollout from the same group} \\
Privileged context (Interactive agents) & \multicolumn{4}{c}{the skill document of the task type} \\
\bottomrule
\end{tabular}
}
\end{table}

LEAD is implemented on top of \texttt{verl}~\citep{verl}. All experiments are conducted on $8\times$H800 GPUs with FSDP, bfloat16 parameters and gradient checkpointing, with $G=8$ on-policy rollouts per query. Specifically:
\begin{itemize}[leftmargin=*, itemsep=2pt, topsep=2pt, parsep=0pt]
\item \textbf{Scientific reasoning.} Each batch samples $32$ questions with $8$ rollouts per question, a maximum prompt length of $2{,}048$ and a maximum response length of $8{,}192$ tokens. Each question is answered in a single turn, and training runs for $10$ epochs over the corresponding training split.
\item \textbf{Interactive agents.} We follow the protocol of SDAR~\citep{lu2026sdar} and train for $150$ steps on each benchmark. For ALFWorld we adopt the training data split of~\citet{feng2025gigpo}, with each batch sampling $16$ tasks and $8$ rollouts per task and a maximum prompt length of $2{,}048$ tokens. For Search we follow the setup of Search-R1~\citep{jin2025searchr1} and use E5 as the retriever; the training data are drawn from NQ and HotpotQA, so that two of the seven evaluation QA datasets are seen during training and the remaining five are not; the reported Search score averages all seven. Each batch samples $128$ tasks with a maximum prompt length of $4{,}096$ tokens. For WebShop, $1{,}000$ tasks are selected for training, with each batch sampling $16$ tasks and $8$ rollouts per task and a maximum prompt length of $4{,}096$ tokens. A single turn emits at most $512$ tokens, and an episode is capped at $50$, $15$ and $4$ turns on ALFWorld, WebShop and Search respectively. In all three environments the privileged context is a skill document retrieved from the SkillBank used by~\citet{lu2026sdar}.
\end{itemize}

More Method-specific hyperparameters are listed in Table~\ref{app:method_para}.

\section{Deferred Proofs}

\subsection{Proof of Proposition 1}
\label{app:proof}

Consider a batch of
\(N\) rollouts
\[
z_i=(x_i,y_i,r_i),
\qquad
i=1,\ldots,N,
\]
where \(x_i\) is the input, \(y_i\) is the sampled rollout, and
\(r_i\) denotes the privileged information associated with
rollout \(i\). Let \(A_i\in\mathbb{R}\) denote the scalar advantage of
rollout \(i\).

For a token \(y_{i,t}\) in rollout \(i\), define the RL log-probability as
\[
u_{i,t}(\theta)
=
\log \pi_\theta
\left(
y_{i,t}\mid x_i,y_{i,<t}
\right),
\]
and define the corresponding privileged log-probability as
\[
s_{i,t}(\theta)
=
\log \pi_\theta
\left(
y_{i,t}\mid x_i,y_{i,<t},r_i
\right).
\]

Their gradients with respect to the model parameters are
\[
v_{i,t}
=
\nabla_\theta u_{i,t}(\theta),
\qquad
h_{i,t}
=
\nabla_\theta s_{i,t}(\theta).
\]

For each rollout \(i\), define its aggregate RL gradient as
\[
V_i
=
\sum_k v_{i,k}.
\]
We consider the following first-order batch policy-gradient update:
\[
\theta^+
=
\theta
+
\eta
\sum_{j=1}^{N}
A_j V_j,
\]
where \(\eta>0\) is the learning rate. Equivalently,
\[
\theta^+
=
\theta
+
\eta
\sum_{j=1}^{N}
\sum_k
A_j v_{j,k}.
\]

For token \(y_{i,t}\), we evaluate the change in the privileged log-probability after
this update:
\[
\Delta_{i,t}
=
s_{i,t}(\theta^+)-s_{i,t}(\theta).
\]

Since \(s_{i,t}(\theta)\) is differentiable with respect to \(\theta\), its
first-order Taylor expansion around \(\theta\) gives
\[
s_{i,t}(\theta^+)
=
s_{i,t}(\theta)
+
\nabla_\theta s_{i,t}(\theta)^\top
(\theta^+-\theta)
+
o\!\left(
\|\theta^+-\theta\|_2
\right).
\]
Using
\[
\nabla_\theta s_{i,t}(\theta)=h_{i,t},
\]
we obtain
\[
\Delta_{i,t}
=
{h_{i,t}}^{\top}
(\theta^+-\theta)
+
o\!\left(
\|\theta^+-\theta\|_2
\right).
\]

Substituting the batch policy-gradient update gives
\[
\begin{aligned}
\Delta_{i,t}
&=
\eta
{h_{i,t}}^{\top}
\sum_{j=1}^{N}
A_jV_j
+
o(\eta)
\\
&=
\eta
\sum_{j=1}^{N}
A_j
{h_{i,t}}^{\top}V_j
+
o(\eta).
\end{aligned}
\]

We now separate the contribution from the same rollout \(i\) from those
of all other rollouts:
\[
\begin{aligned}
\Delta_{i,t}
&=
\eta A_i {h_{i,t}}^{\top}V_i
+
\eta
\sum_{j\neq i}
A_j {h_{i,t}}^{\top}V_j
+
o(\eta).
\end{aligned}
\]

The first term contains the aggregate RL gradient generated by rollout
\(i\) itself and it therefore represents the contribution of the target rollout's own training signal to its post-update privileged log-probability. We refer to this term as the self-induced component.
The second term contains only the training signals generated by other rollouts. We refer to this term as the cross-scoring component.

Therefore,
\[
\Delta_{i,t}
=
\underbrace{
\eta A_i {h_{i,t}}^{\top}V_i
}_{\text{self-induced}}
+
\underbrace{
\eta
\sum_{j\neq i}
A_j {h_{i,t}}^{\top}V_j
}_{\text{cross-induced}}
+
o(\eta),
\]
which proves the proposition.

\begin{table}[h]
\centering
\scriptsize
\caption{Curated cases of Physics domain.}
\label{app_tab:cases_phys}
\resizebox{\textwidth}{!}{%
\begin{tabular}{cp{5.2cm}p{1.1cm}p{3.3cm}}
\toprule
Domain / split  & Original item & Original label & Curation \\
\midrule
Physics / train & {\ttfamily Question: A current-carrying helix has 518 turns with a radius of 7 cm and a current of 5 \newline A: Calculate the magnetic field at a point that is located on the axis of the helix and is 74 cm away from its center. Additionally, please select the option from the given choices that you believe is closest to the correct answer! (Please note that the following content involves constants might be used in the calculations : \textbackslash{}mu\_0 = 1.26 * 10\^{}(-6) V$\cdot$s/(A$\cdot$m) \newline B: 1.94 x 10\^{}(-5) Tesla \newline C: 8.65 x 10\^{}(-5) Tesla \newline D: 4.53 x 10\^{}(-5) Tesla} & A & \emph{relabel}: ground truth A $\rightarrow$ B \\
\midrule
Physics / test & {\ttfamily Question: Determine the  magnetic field strength at a distance of 1  cm from a straight wire carrying a current of 2 \newline A: The wire has a length of 2 cm, and the current flows in the direction perpendicular to the wire. Additionally, please select the option from the given choices that you believe is closest to the correct answer! (Please note that the following content involves constants might be used in the calculations : \textbackslash{}mu\_0 = 1.26 * 10\^{}(-6) V$\cdot$s/(A$\cdot$m) \newline B: 66.43 x 10\^{}(-6) Tesla \newline C: 80.87 x 10\^{}(-6) Tesla \newline D: 40.0 x 10\^{}(-6) Tesla} & C & \emph{relabel}: ground truth C $\rightarrow$ D \\
\midrule
Physics / train  & {\ttfamily Question: What is the density of a plasma with a temperature of 11 electron volts and an electron number density of 2.4e+19 m\^{}-3? Use the Saha equation to determine the ionization state of the plasma assuming it consists of only hydrogen atoms. Please note that the following content involves constants might be used in the calculations : Vacuum tolerance $\varepsilon$ = 8.85e-12 ; Boltzmann constant k = 1.38e-23 ; electronic mass m\_e = 9.11e-31 kg ; e = 1.6e-19 C ; proton mass m\_p = 1.673e-27 kg ; neutral hydrogen atom mass m\_h = 1.673e-27 kg. Additionally, please select the option from the given choices that you believe is closest to the correct answer! \newline A: 4.02e-27 kg/m\^{}3 \newline B: 1.28e-27 kg/m\^{}3 \newline C: 8.26e-27 kg/m\^{}3 \newline D: 8.2e-27 kg/m\^{}3} & A & \emph{rewrite}: option A rewritten to \texttt{4.02e-08 kg/m\^{}3} and ground truth stays A \\
\bottomrule
\end{tabular}
}
\end{table}

\begin{table}[h]
\centering
\scriptsize
\caption{Curated cases of Chemistry and Materials Science domains.}
\label{app_tab:cases_other}
\begin{tabular}{cp{5.2cm}p{1.1cm}p{3.3cm}}
\toprule
Domain / split & Original item & Original label & Curation \\
\midrule
Chemistry / train & {\ttfamily What is the molar weight (g/mol) of the molecule with the canonical SMILES representation 'C1CC(NC1)C(=O)O'? \newline A: 118.150 \newline B: 152.140 \newline C: 114.150 \newline D: 115.150} &  A & \emph{relabel}: A $\rightarrow$ D \\
\midrule
Chemistry / train & {\ttfamily What is the molar weight (g/mol) of the molecule with the canonical SMILES representation 'C1CC(NC1)C(=O)O'? \newline A: 122.170 \newline B: 145.200 \newline C: 114.150 \newline D: 182.020} & A & \emph{rewrite}: option A rewritten to \texttt{115.130} and ground truth stays A \\
\midrule
Material / train & {\ttfamily Given the following data on dilute solute elements and their diffusion barriers in different host elements, compare these values and answer the question. The diffusion barrier of the dilute solute elements (in eV) is calculated using DFT methods. Choose the combination with the highest diffusion rate. \newline Given data: \newline Solute Element 1: Fe, Host Element 1: Pd, Diffusion Barrier 1: -0.14447 eV \newline Solute Element 2: Fe, Host Element 2: Pt, Diffusion Barrier 2: -0.00302 eV \newline Solute Element 3: Fe, Host Element 3: Sc, Diffusion Barrier 3: -0.28547 eV \newline Solute Element 4: Fe, Host Element 4: Ta, Diffusion Barrier 4: -0.19922 eV \newline A: Fe in Pd \newline B: Fe in Pt \newline C: Fe in Sc \newline D: Fe in Ta} & B & \emph{relabel}: B $\rightarrow$ C \\
\bottomrule
\end{tabular}
\end{table}

\end{document}

%% file: math_commands.tex
\usepackage{amsmath,amsfonts,bm}

\def\eqref#1{equation~\ref{#1}}
\def\1{\bm{1}}

\DeclareMathAlphabet{\mathsfit}{\encodingdefault}{\sfdefault}{m}{sl}
\SetMathAlphabet{\mathsfit}{bold}{\encodingdefault}{\sfdefault}{bx}{n}